\documentclass[10pt,twocolumn,letterpaper]{article}

\usepackage[pagenumbers]{cvpr} 

\graphicspath{{sec/}}

\ifdefined\pdfinclusioncopyfonts
\fi

\usepackage[T1]{fontenc}
\usepackage{microtype}
\usepackage{multirow}
\usepackage{tikz}
\usetikzlibrary{positioning,arrows.meta,fit,calc,backgrounds}

\renewcommand{\paragraph}[1]{\vspace{.2em}\noindent\textbf{#1}\hspace{0.45em}}

\makeatletter
\def\egosection{\@startsection{section}{1}{\z@}{-1.5ex plus -0.4ex minus -0.2ex}{0.7ex plus 0.2ex}{\large\bf}}
\def\egosubsection{\@startsection{subsection}{2}{\z@}{-1.2ex plus -0.3ex minus -0.2ex}{0.45ex plus 0.15ex}{\elvbf}}
\def\egosubsubsection{\@startsection{subsubsection}{3}{\z@}{-0.9ex plus -0.2ex minus -0.1ex}{0.3ex plus 0.1ex}{\tenbf}}
\def\egosect#1{\egosection{\texorpdfstring{\hskip -1em.~}{}#1}}
\def\egosubsect#1{\egosubsection{\texorpdfstring{\hskip -1em.~}{}#1}}
\def\egosubsubsect#1{\egosubsubsection{\texorpdfstring{\hskip -1em.~}{}#1}}
\renewcommand{\section}{\@ifstar{\egosection*}{\egosect}}
\renewcommand{\subsection}{\@ifstar{\egosubsection*}{\egosubsect}}
\renewcommand{\subsubsection}{\@ifstar{\egosubsubsection*}{\egosubsubsect}}
\makeatother

\AtBeginDocument{%
  \setlength{\abovedisplayskip}{4pt plus 1pt minus 1pt}%
  \setlength{\belowdisplayskip}{4pt plus 1pt minus 1pt}%
  \setlength{\abovedisplayshortskip}{2pt plus 1pt}%
  \setlength{\belowdisplayshortskip}{2pt plus 1pt}%
}

\definecolor{bpOurs}{RGB}{0,114,178}
\definecolor{bpPrior}{RGB}{213,94,0}
\definecolor{bpData}{RGB}{0,158,115}
\definecolor{bpGray}{RGB}{110,110,110}
\definecolor{bpLight}{RGB}{240,242,245}
\colorlet{bpOursFill}{bpOurs!14}
\colorlet{bpDataFill}{bpData!14}

\tikzset{
  bp/base/.style   ={draw=black!70, line width=0.5pt, rounded corners=2pt, align=center,
                     font=\sffamily\scriptsize, minimum height=0.40cm, inner sep=2.3pt},
  bp/frozen/.style ={bp/base, fill=bpLight, draw=bpGray, text=black!85},
  bp/ours/.style   ={bp/base, fill=bpOursFill, draw=bpOurs, line width=0.65pt, text=black},
  bp/data/.style   ={bp/base, fill=bpDataFill, draw=bpData, text=black},
  bp/flow/.style   ={-{Latex[length=1.4mm]}, line width=0.6pt, black!75},
}

\definecolor{cvprblue}{rgb}{0.21,0.49,0.74}
\usepackage[pagebackref,breaklinks,colorlinks,allcolors=cvprblue]{hyperref}

\title{Ego2X: Contact-aware Transfer from Egocentric Human Data to Any Robot}

\author{
  Xinke Lyu$^{1}$ \quad Bowen Yang$^{2}$ \quad Zhicheng Wang$^{3}$ \\[0.5em]
  $^{1}$Beijing University of Posts and Telecommunications \\
  $^{2}$Institute of Automation, Chinese Academy of Sciences
  $^{3}$PowerTech(Mite.)
}

\begin{document}

\twocolumn[{%
\renewcommand\twocolumn[1][]{#1}%
\maketitle
\vspace{-2.2em}
\begin{center}
  \centering
  \includegraphics[width=0.91\linewidth]{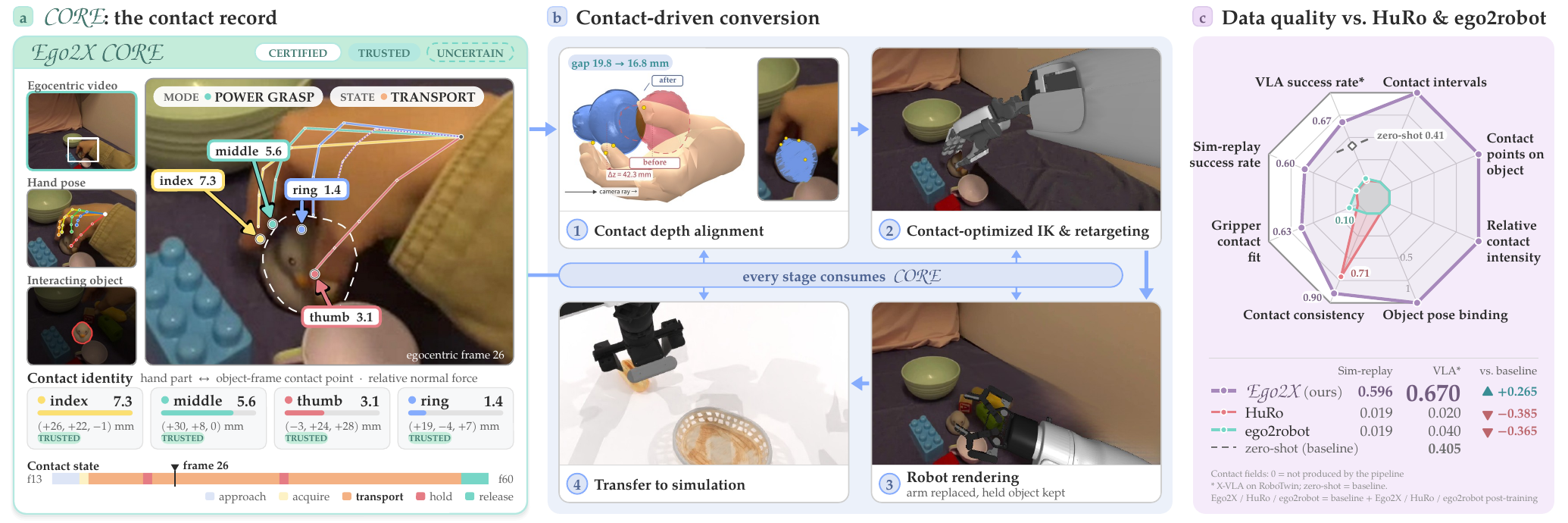}
  \captionof{figure}{\textbf{Ego2X transfers object-frame contacts rather than hand motion.} From monocular video, Ego2X \textbf{(a)} recovers object-frame CORE contact records, \textbf{(b)} drives decoupled depth alignment, cross-morphology contact IK, de-skinned robot compositing, and simulation transfer, and \textbf{(c)} outperforms prior pipelines across exported contact fields, closed-loop simulation conversion ($59.6\%$ vs.\ $1.9\%$), and $200$-scene X-VLA success rate ($67.0\%$ vs.\ $40.5\%$ zero-shot).}
  \label{fig:teaser}
\end{center}
\vspace{-0.4em}
}]

\begingroup
\renewcommand{\thefootnote}{\fnsymbol{footnote}}
\footnotetext[1]{This work was done when Xinke and Bowen were interns at PowerTech.}
\endgroup

\begin{abstract}
Embodied models depend on real interaction data, yet collecting it on physical robots is slow and labor-intensive. Recent ego-to-robot pipelines retarget first-person view human videos to robot data via transferring hand and arm motion. They judge fidelity mainly by visual appearance. However, trajectories often fail in terms of physics: fingers hover beside objects, and grippers close on empty space. We argue that how the object moves and the contacts that drive the motion count. Thus, we present Ego2X, an object-centric framework that converts monocular egocentric video into physically executable robot data across dexterous hands and parallel grippers. Ego2X centers on CORE (Contact-centric Object-interaction REpresentation), which records each contact on the object by its surface anchor, as well as relative force, grasp mode, and interaction state. Throughout the pipeline, CORE corrects hand--object drift via object-relative constraints, steers inverse kinematics toward contacts the robot must realize, and preserves held objects during arm erasure and robot compositing. On HOT3D, contact-driven correction reduces in-contact hand depth error by $33.7\%$, while on EgoDex, contact-driven retargeting realizes $62.9\%$ of annotated contacts, surpassing the official HuRo pipeline ($27.9\%$). In RoboTwin~2.0, post-training X-VLA on Ego2X episodes alone, without simulator demonstrations, raises task success rate from $40.5\%$ zero-shot to $67.0\%$ across $200$ paired scenes. We will publicly release our conversion pipeline and CORE-annotated dataset for dexterous hands and parallel grippers.
\end{abstract}

\section{Introduction}
\label{sec:intro}

Vision-language-action policies and robot world models depend on physical interaction data. Teleoperated demonstrations are slow to collect and tied to specific hardware~\cite{aloha,droiddataset}, whereas egocentric human videos capture diverse manipulation skills at scale~\cite{egodex,hot3d,ego4d}. Converting human video into robot observations and actions provides a direct route to scalable embodied learning.

Recent ego-to-robot pipelines perform this conversion by retargeting 3D human hand kinematics to robot joints~\cite{huro,ego2robot,phantom,egomimic}. Phantom~\cite{phantom} composites rendered arms over inpainted hands, EgoMimic~\cite{egomimic} aligns human and robot action spaces, and HuRo~\cite{huro} and Ego2Robot~\cite{ego2robot} map reconstructed 3D hand trajectories via inverse kinematics. Their transferred unit is hand motion. The monocular hand estimates drift by centimeters along the camera ray, whereas stable manipulation demands millimeter precision on the object surface. Consequently, retargeted fingers hover beside objects, and parallel grippers close on empty space.


Recovering and transferring object-frame contact from monocular RGB video poses three coupled challenges. First, monocular hand and object estimators exhibit depth and scale drift along the viewing ray. Therefore, using the hand to fix the object scale and that object to correct the hand depth couples their errors in a circular loop. Second, contacts occur on occluded surfaces and span timing, surface geometry, and force distribution that raw images do not directly expose. Third, the recovered contacts must drive robots with distinct kinematics and contact surfaces while replacing the human arm in video without erasing the held object.

\begin{figure*}[t]
  \centering
  \includegraphics[width=\textwidth]{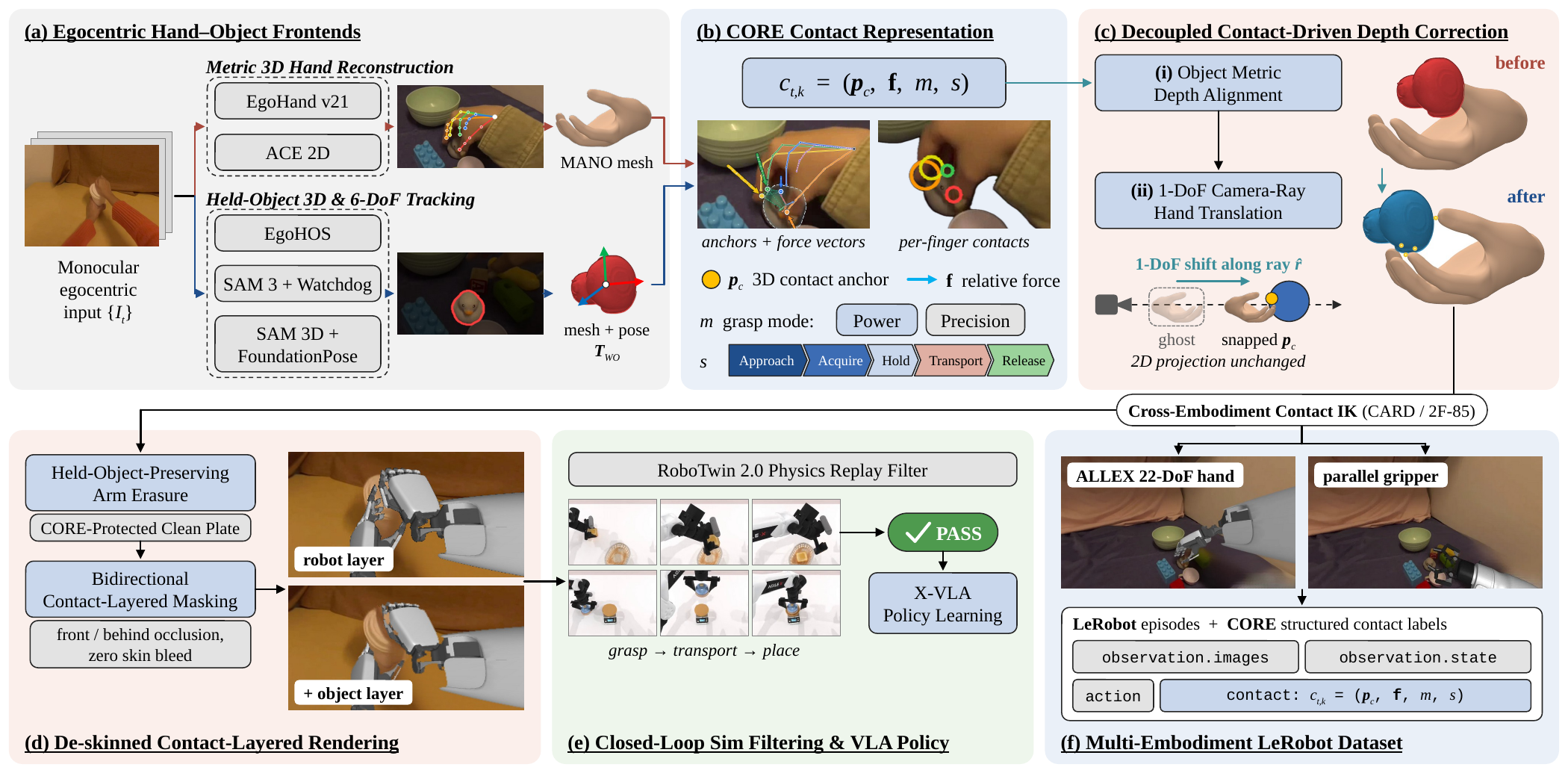}
  \caption{\textbf{Ego2X pipeline and CORE overview.} From monocular video $\{I_t\}$, Ego2X reconstructs metric hands and signed object surfaces $\phi_o$ (\cref{sec:geometry}) to extract object-frame CORE records $\mathbf{c}_{t,k} = (\mathbf{p}_{t,k}, \mathbf{f}_{t,k}, m_{t,k}, s_{t,k})$ (\cref{sec:core}), driving decoupled depth correction, cross-morphology retargeting, de-skinned compositing, and closed-loop simulation verification (\cref{sec:recovery,sec:embodiment,sec:sim}).}
  \label{fig:pipeline}
  \vspace{-0.4em}
\end{figure*}

We present \textbf{Ego2X}, an object-centric framework that converts monocular egocentric video into physically executable robot data across dexterous hands and parallel grippers (\cref{fig:teaser}). Ego2X centers on \textbf{CORE} (Contact-centric Object-interaction REpresentation), which records each contact in the object frame by its surface anchor, outward normal, push direction, relative force, grasp mode, and interaction state (\cref{fig:pipeline}). Recovered from tactile timing and camera-ray surface intersection, CORE unifies every stage of the conversion pipeline. To resolve ray-depth drift without circular scale coupling, Ego2X diagnoses object pose reliability and decouples depth estimation. Furthermore, it fuses hand scale with multi-view ray triangulation on the object path while shifting the hand along the camera ray against hand-ruler-free multi-view object geometry. To bridge morphologies, a URDF compiler maps robot collision geometry into mode-compatible contact pads, and contact-driven inverse kinematics pulls dexterous pads onto CORE anchors or synthesizes antipodal object-frame grasps for parallel grippers. Finally, CORE interaction states guide object-protected arm erasure and contact-edge de-skinning during video compositing, while closed-loop simulation replay verifies physical task completion.

Across HOT3D~\cite{hot3d}, EgoDex~\cite{egodex}, and RoboTwin~2.0~\cite{robotwin2}, Ego2X substantially improves geometric accuracy, physical contact realization, and downstream policy performance. On HOT3D, contact-driven depth correction cuts in-contact hand ray-depth error by $33.7\%$ under tracked object poses and lowers monocular object translation error by $30.4$--$47.3\,\mathrm{mm}$ under estimated poses. On EgoDex, contact-driven retargeting realizes $62.9\%$ of annotated contacts for dexterous hands compared to $27.9\%$ for the official HuRo pipeline~\cite{huro}, and achieves $63.2\%$ physical contact fit for parallel grippers compared to $9.7\%$ for Ego2Robot~\cite{ego2robot}. In RoboTwin~2.0, post-training X-VLA~\cite{xvla} on Ego2X episodes alone, without simulator demonstrations, raises task success rate from $40.5\%$ zero-shot to $67.0\%$ across $200$ paired scenes.

Our contributions are threefold:
\begin{itemize}\setlength{\itemsep}{1pt}\setlength{\parskip}{0pt}\setlength{\topsep}{2pt}
  \item \textbf{Contact-centric Object-interaction REpresentation (CORE)}: We identify that object interaction dynamics constitute the primary transferable value in egocentric videos. To capture this, we formulate an object-centric conversion paradigm that explicitly models contact points, forces, grasp modes, and interaction states within the object coordinate frame (\cref{sec:core}).
  \item \textbf{CORE-driven Ego2X Pipeline}: Leveraging our contact-aware representation, we develop the Ego2X framework to bridge the human-to-robot embodiment gap (\cref{sec:method}). Specifically, it integrates contact-driven hand--object depth correction (\cref{sec:hand-refine}), cross-morphology contact retargeting (\cref{sec:contact-ik,sec:embodiment}), and contact-guided layered video rendering (\cref{sec:render}).
  \item \textbf{Egocentric Video Post-training for VLAs}: As far as we know, we are the first to exploit post-training of embodied models using egocentric videos, specifically, with object interaction and contact priors from egocentric videos to synthesize robot trajectories (\cref{sec:sim}). Without any expert demonstrations in simulation, our converted data effectively post-trains downstream Vision-Language-Action (VLA) models and substantially improves their task success rates.
\end{itemize}

\section{Related Work}
\label{sec:related}

\paragraph{Learning manipulation from human videos.}
Human videos support robot manipulation learning primarily through policy alignment or embodiment conversion. Policy alignment methods extract latent plans~\cite{mimicplay}, align action spaces~\cite{egomimic}, imitate object-centric motions on fixed humanoids~\cite{okami}, or pretrain VLA models~\cite{egovla}. Conversely, embodiment conversion pipelines composite rendered arms over inpainted hands~\cite{phantom}, retarget 3D hand trajectories to robot joints~\cite{huro,ego2robot}, or edit videos generatively~\cite{roboedit}, with recent benchmarks evaluating policy transfer~\cite{h2rbench}. Unlike pipelines that transfer free-space hand motion or single-robot wrist poses, Ego2X compiles object-frame contacts via CORE across diverse morphologies.

\paragraph{Hand--object reconstruction and contact.}
Monocular hand estimators~\cite{wilor,egohand,hawor}, 3D object models and trackers~\cite{sam3d,foundationpose}, and hand--object benchmarks~\cite{dexycb,hot3d,egodex} provide 3D geometric priors from RGB inputs. Prior works model hand--object contact via 2D interaction regions~\cite{doh100,egohos}, visual tactile cues~\cite{pressurevision,egotac}, mocap contact maps~\cite{grab}, grasp taxonomies~\cite{feix2016grasp}, and contact-guided mesh fitting~\cite{contactopt,c2dex}. Building on these priors, Ego2X fuses tactile timing, camera-ray intersection, and two-part load priors in the object frame to correct ray drift without circular scale coupling.

\paragraph{Cross-embodiment retargeting.}
Kinematic retargeting typically maps human keypoints to dexterous hands~\cite{dexmv,anyteleop} or character surfaces~\cite{contactretarget}. In parallel, simulation trackers apply residual RL or curricula around ground-truth 3D object trajectories~\cite{maniptrans,dexmachina}, while cross-embodiment policies~\cite{crossformer,xvla} and grasp generators~\cite{drograsp} bridge distinct morphologies. Rather than training per-clip RL trackers on mocap trajectories, Ego2X compiles URDFs into mode-compatible pads and solves object-frame targets from monocular RGB for both dexterous hands and parallel grippers.

\paragraph{Robot video synthesis and physical verification.}
In visual synthesis, robot-to-robot repainting~\cite{mirage,roviaug} swaps manipulators in robot logs, and generative editing inserts robot arms into human videos~\cite{roboedit}. Meanwhile, real-to-sim reconstruction~\cite{rialto}, bimanual benchmarks~\cite{robotwin2}, and physics-in-the-loop optimization~\cite{hsimul3r} evaluate dynamic feasibility in simulation. Ego2X unifies both directions by coupling de-skinned compositing with closed-loop physical replay around one shared contact state.
\section{Ego2X}
\label{sec:method}

Given a monocular egocentric video $\{I_t\}_{t=1}^{T}$ and a target robot URDF, Ego2X reconstructs metric hands and signed object surfaces (\cref{sec:geometry}) and extracts object-frame CORE records (\cref{sec:core}). From these records, it outputs a verified LeRobot~\cite{lerobot} episode comprising robot actions, training masks, and an object-preserving video (\cref{fig:pipeline,sec:recovery,sec:embodiment}).

\subsection{Hand and Object Geometry}
\label{sec:geometry}

\noindent\textbf{Camera and interaction masks.}
We reconstruct hands and objects in a gravity-aligned metric world frame. Ego2X uses calibrated intrinsics $\mathbf{K}$ and camera poses $\mathbf{T}_{WC}$ when available~\cite{egodex,hot3d}. For uncalibrated video, DROID-calib~\cite{droidcalib} estimates $\mathbf{K}$, DROID-SLAM~\cite{droidslam} tracks $\mathbf{T}_{WC}$, MoGe-2~\cite{moge2} recovers metric scale, and GeoCalib~\cite{geocalib} aligns gravity. Within tactile contact windows, EgoHOS~\cite{egohos} hand and active-object masks seed SAM~3~\cite{sam3} video propagation with backward re-anchoring across occlusions.

\noindent\textbf{Metric hand and object surfaces.}
WiLoR~\cite{wilor} detects 2D hand boxes that ACE-Ego-Hand~\cite{aceegohand} completes across occlusions, from which EgoHand~\cite{egohand} predicts metric 3D hand skeletons. We fit a per-frame MANO~\cite{mano} mesh with clip-consistent shape and partition its surface into 24 anatomical regions (\cref{tab:modes}). For objects, SAM~3D~\cite{sam3d} reconstructs a canonical mesh with an exact non-convex signed distance field (SDF) $\phi_o$. FoundationPose~\cite{foundationpose} then tracks metric 6-DoF poses $\mathbf{T}^{t}_{WO}$, regularized via $\mathrm{SO}(3)$ projection and gravity-aligned symmetry resolution.

\subsection{CORE: Contact-Centric Object-Interaction Representation}
\label{sec:core}

Raw 3D hand trajectories vary across embodiments and drift under monocular depth uncertainty. CORE (\cref{fig:pipeline}) instead represents manipulation in the object frame as per-frame four-tuple contact records:
\begin{equation}
  \mathbf{c}_{t,k} = \bigl(\mathbf{p}_{t,k},\ \mathbf{f}_{t,k},\ m_{t,k},\ s_{t,k}\bigr),
  \label{eq:core}
\end{equation}
paired with an evidence tier $\ell_k\in\{\textsc{Certified},\textsc{Trusted},\textsc{Hypothesis},\textsc{Uncertain}\}$. Here, $\mathbf{p}_{t,k} = (r_k,\ \mathbf{p}^{O}_{k},\ \mathbf{n}^{O}_{k})$ stores the contacting MANO region $r_k$, object-frame surface anchor $\mathbf{p}^{O}_{k}\in\mathbb{R}^3$, and outward unit normal $\mathbf{n}^{O}_{k}\in\mathbb{S}^2$. The remaining terms specify a two-part force prior $\mathbf{f}_{t,k} = (\rho_{t,k},\ f^{\mathrm{lb}}_{t,k},\ \hat{\mathbf{d}}_{t,k})$, the grasp mode $m_{t,k}$ (\cref{tab:modes}), and the interaction state $s_{t,k}\in\{\textsc{Approach},\textsc{Acquire},\textsc{Hold},\textsc{Transport},\textsc{Release}\}$.

\subsubsection{Contact Points}
\label{sec:anchors}

Monocular hand depth drifts along the viewing ray, whereas 2D projections remain well-localized. We therefore anchor contacts by intersecting camera rays with the object surface. EgoTac~\cite{egotac} detects contact timing from tactile appearance ($\tau_{t,v}>0.5$). For each active hand vertex $v$, we cast a ray $\mathbf{x}_{t,v}(\lambda)=\mathbf{o}_t+\lambda\,\mathbf{u}_{t,v}$ from camera center $\mathbf{o}_t$ along pixel ray $\mathbf{u}_{t,v}$ to the first zero-crossing of $\phi_o$:
\begin{equation}
  \lambda^{*} = \min\bigl\{\lambda>0 : \phi_o\bigl(\mathbf{T}^{t}_{OW}(\mathbf{o}_t+\lambda\,\mathbf{u}_{t,v})\bigr)=0\bigr\},
  \label{eq:ray}
\end{equation}
retaining hits whose outward normal $\mathbf{n}^{O}_{k}$ opposes the palmar normal. We cluster object-frame hits per region and phase into medoid anchors $\mathbf{p}^{O}_{k}$. Medoid bridging connects intervals, grasp modes, and states across transient occlusions, while CORE's field-selective readout retains observed or rigidly carried camera-ray hits for 3D anchor coordinates rather than interpolated bridge midpoints. Each anchor is graded into \textsc{Certified}, \textsc{Trusted}, \textsc{Hypothesis}, or \textsc{Uncertain} based on palmar orientation, adjacent-phalanx non-penetration, rigid-motion agreement, and 2D occlusion ordering. \textsc{Hypothesis} and \textsc{Uncertain} slots preserve temporal continuity across $(m_{t,k},s_{t,k})$, whereas only \textsc{Certified} and \textsc{Trusted} anchors constrain downstream depth refinement and inverse kinematics.

\subsubsection{Contact Modes and States}
\label{sec:modes}

\begin{table}[t]
  \centering
  \scriptsize
  \renewcommand{\arraystretch}{0.94}
  \setlength{\tabcolsep}{3.0pt}
  \caption{\textbf{Contact modes in CORE.} Active MANO regions determine the grasp mode and the robot pads that realize it.}
  \label{tab:modes}
  \begin{tabular*}{\columnwidth}{@{\extracolsep{\fill}}lll@{}}
    \toprule
    Mode & Regions in contact & Robot pads used \\
    \midrule
    Palm carry         & palm, ${\le}1$ finger, object above palm  & palm + guiding fingers \\
    Power grasp        & palm + ${\ge}3$ finger regions            & palm + all phalanges \\
    Intermediate grasp & palm + 1--2 finger regions                & palm + all phalanges \\
    Fingertip hold     & mid/distal phalanges (${\ge}2$), no palm  & middle \& distal pads \\
    Precision pinch    & thumb \& index phalanges, no palm         & middle \& distal pads \\
    Hook grasp         & proximal/middle phalanges, no palm        & proximal \& middle pads \\
    Unknown            & tactile contact, no 3D support            & unconstrained \\
    \bottomrule
  \end{tabular*}
  \vspace{-0.5em}
\end{table}

We classify each contact interval into one of seven grasp modes (\cref{tab:modes}) based on the palmar regions contacting $\phi_o$. Contact boundaries and object motion then partition the manipulation sequence into the five interaction states $s_{t,k}$, synchronizing gripper actuation, object pixel protection, and simulation waypoints.

\subsubsection{Contact Force Prior}
\label{sec:forces}

Without tactile sensors, monocular RGB cannot uniquely determine internal grasp forces in Newtons. CORE therefore models $\mathbf{f}_{t,k} = (\rho_{t,k},\ f^{\mathrm{lb}}_{t,k},\ \hat{\mathbf{d}}_{t,k})$ as a structured two-part force prior that combines visual part-load estimation with kinematic support-load scaling.

\noindent\textbf{Visual relative part-load distribution.}
EgoTac~\cite{egotac} predicts per-region contact probabilities and uncalibrated intensities from hand--object appearance. Normalizing intensities across active regions yields relative load shares $\rho_{t,k}\in[0,1]$ ($\sum_k \rho_{t,k}=1$) and inward push directions $\hat{\mathbf{d}}_{t,k}\in\mathbb{S}^2$ opposing $\mathbf{n}^{O}_{k}$ within the local friction cone.

\noindent\textbf{Kinematic normal and support load prior.}
During hand-borne \textsc{Hold} and \textsc{Transport} frames without external support, Gaussian-smoothed ($\sigma=1.0$) object center-of-mass trajectories yield the required specific support $\mathbf{s}_{\mathrm{req},t} = \mathbf{a}_{\mathrm{COM},t}-\mathbf{g}$. We estimate object mass $m_{\mathrm{band}}=\varrho_{\mathrm{band}}V$ from watertight mesh volume $V$ across reference density bands $\varrho_{\mathrm{band}}\in\{0.5, 1.0, 1.5\}\,\mathrm{g/mL}$ (default $1.0\,\mathrm{g/mL}$, or mass-free $\|\mathbf{s}_{\mathrm{req},t}\|$ on open meshes). Allocating this load via EgoTac hand share $w^{\mathrm{hand}}_{t}$ and region share $\rho_{t,k}$ defines the lower-bound normal support prior:
\begin{equation}
  f^{\mathrm{lb}}_{t,k} = m_{\mathrm{band}}\,\|\mathbf{a}_{\mathrm{COM},t}-\mathbf{g}\|\,w^{\mathrm{hand}}_{t}\,\rho_{t,k}.
  \label{eq:force}
\end{equation}
Downstream, $\rho_{t,k}$ and $\hat{\mathbf{d}}_{t,k}$ weight anchor attraction and pad-normal alignment in dexterous IK (\cref{eq:ik}), while $f^{\mathrm{lb}}_{t,k}$ records physically grounded support-load scales.

\subsection{CORE-Driven Interaction Recovery}
\label{sec:recovery}

\subsubsection{Credibility-Aware Decoupled Depth Correction}
\label{sec:hand-refine}

Jointly scaling the object from the hand while pulling the hand toward the object induces circular ray-depth drift. We therefore score pose credibility and decouple the object and hand refinement paths.

\noindent\textbf{Pose credibility.}
A monotonic GBDT trained via leave-one-clip-out (LOCO) cross-validation maps 29 GT-free contour, multi-view, depth, and temporal features to identify usable object poses ($t_{\mathrm{err}}<30\,\mathrm{mm}$, $\theta_{\mathrm{sym}}<15^\circ$). Under monocular scale ambiguity, isotonic-calibrated single-frame confidence peaks at $0.557$, causing hard single-frame gating to starve coverage. Instead, we decouple the two paths without gating the hand per frame.

\noindent\textbf{Object and hand paths.}
On the object path, $\mathbf{T}^{t}_{WO}$ aligns to \textsc{Certified} contacts and fuses hand scale with windowed multi-view SIFT ray triangulation, updating only the exported object trajectory. On the hand path, we shift the hand by $\delta\in[-150,+50]$\,mm along its root camera ray by maximizing a Cauchy objective over \textsc{Certified} and \textsc{Trusted} vertex--anchor pairs $(v,a)$ ($\ge 3$ inliers):
\begin{equation}
  S(\delta) = \textstyle\sum_{(v,a)} \frac{w_{a}}{1 + r_{v,a}(\delta)^2/\sigma_r^2},
  \qquad \sigma_r = 15\,\mathrm{mm}.
  \label{eq:dr}
\end{equation}
Residuals $r_{v,a}(\delta)$ anchor to $\mathbf{T}^{t}_{WO}$ when external tracking is available, and strictly to hand-ruler-free multi-view object triangulation under estimated poses ($\delta\equiv 0$ below $15\,\mathrm{mm}$ parallax). We then linearly interpolate 15-frame inner-edge shifts across each contact interval while preserving 2D projections.

\subsubsection{Contact-Driven Inverse Kinematics}
\label{sec:contact-ik}

\noindent\textbf{Dexterous hands.}
From a clip-fixed torso base~\cite{huro}, we solve per-frame robot joint angles $\mathbf{q}$ by minimizing
\begin{equation}
\begin{aligned}[b]
  & E_t(\mathbf{q}) = E_{\mathrm{shape}} + E_{\mathrm{obj}} + E_{\mathrm{self}} + E_{\mathrm{mask}} + E_{\mathrm{temp}} \\
  & + \textstyle\sum_a w_a \bigl(\|\mathbf{r}_\sigma(\boldsymbol{\delta}_a)\|^2 + \gamma_a \|\mathbf{n}_{j(a)}(\mathbf{q})\!+\!\mathbf{n}^{W}_{t,a}\|^2 \\
  & \qquad\qquad + \gamma_a g_a \|\mathbf{n}_{j(a)}(\mathbf{q})\!-\!\hat{\mathbf{d}}_{t,a}\|^2\bigr),
\end{aligned}
\label{eq:ik}
\end{equation}
where $\boldsymbol{\delta}_a = \mathbf{p}_{j(a)}(\mathbf{q})-\mathbf{p}^{W}_{t,a}$ is the offset from mode-compatible robot pad $j(a)$ (\cref{tab:modes}) to world anchor $\mathbf{p}^{W}_{t,a}$, $\mathbf{r}_\sigma(\boldsymbol{\delta}) = \boldsymbol{\delta}/\sqrt{1+\|\boldsymbol{\delta}\|^2/\sigma^2}$ is the Geman--McClure residual, and $\gamma_a = 1/\sqrt{1+\|\boldsymbol{\delta}_a\|^2/\sigma^2}$ is a proximity weight. Weight $w_a$ combines evidence tier, state, and relative load $\rho_{t,a}$. As pad $j(a)$ approaches $\mathbf{p}^{W}_{t,a}$, $\gamma_a$ aligns its normal with $-\mathbf{n}^{W}_{t,a}$ and steers it toward $\hat{\mathbf{d}}_{t,a}$ when friction indicator $g_a$ is active. Far from $\phi_o$, $\mathbf{r}_\sigma$ saturates so unreachable anchors release smoothly. The regularizers enforce human-like posture ($E_{\mathrm{shape}}$), object non-penetration ($E_{\mathrm{obj}}$), self-collision avoidance ($E_{\mathrm{self}}$), arm silhouette overlap ($E_{\mathrm{mask}}$), and temporal smoothness ($E_{\mathrm{temp}}$).

\noindent\textbf{Parallel grippers.}
For two-jaw grippers, we search the solid-filled object SDF along closure axes opposing active anchor normals. We identify antipodal contact pairs matching the jaw aperture while keeping non-pad links outside $\phi_o$, and query candidates in tier order to lock one object-frame grasp $\mathbf{T}_{OG}$ per contact interval. Using a clip-fixed table-edge dual-arm mount, states $s_{t,k}$ schedule pre-grasp approach, jaw closure in \textsc{Acquire}, rigid binding $\mathbf{T}_{WG}^{t} = \mathrm{med}_3(\mathbf{T}_{WO}^{t})\,\mathbf{T}_{OG}$ (with a 3-frame temporal median filter $\mathrm{med}_3(\cdot)$) during \textsc{Hold} and \textsc{Transport}, and opening in \textsc{Release}.

\noindent\textbf{Trajectory consistency.}
At a fixed wrist pose, Viterbi dynamic programming over the 7-DoF elbow swivel angle aligns the projected robot arm with the human arm mask. Banded least-squares retiming then enforces joint position and velocity limits ($\le 7.5$\,rad/s) with priority weighting on contact frames.

\subsubsection{Contact-Aware Erasure and Compositing}
\label{sec:render}

\noindent\textbf{Object-protected erasure.}
Inpainting the full hand and arm mask during manipulation erases held objects. During \textsc{Acquire}, \textsc{Hold}, and \textsc{Transport}, we therefore subtract a protected object zone $K_5\setminus H_{+2}$ (the $5$\,px dilated object mask $M$ minus the $2$\,px dilated hand mask $H$ and MANO hull) from the SAM~3 erasure mask before running ProPainter~\cite{propainter}.

\noindent\textbf{Contact-edge de-skinning.}
To eliminate human skin bleed along occluded hand--object boundaries without noisy per-pixel depth ordering, we retain original object pixels during contact within the de-skinned mask:
\begin{equation}
  K_{\mathrm{clean}} = M_{\mathrm{core}} \cup \bigl((M\setminus M_{\mathrm{core}})\setminus S_{\mathrm{rim}}\bigr) \cup \bigl((K_5\setminus M)\setminus H_{+2}\bigr),
  \label{eq:clean}
\end{equation}
where $M_{\mathrm{core}}$ is the eroded object interior. Within the boundary rim $M\setminus M_{\mathrm{core}}$, $S_{\mathrm{rim}}$ strips hand-overlapping pixels whose Lab color is closer to an interval skin prototype ($H\setminus K_5$) than to an interior object prototype ($M_{\mathrm{core}}\setminus H$). Stripped rim pixels expose the rendered robot inside its silhouette and the inpainted background elsewhere, composited with a $3$\,px inward feather.

\subsection{URDF Compilation and Simulation Verification}
\label{sec:embodiment}\label{sec:sim}

To generalize without per-robot tuning, our URDF compiler samples link collision meshes into phalanx/palm or opposed-jaw pad pools with normals, pinch midpoints, aperture maps, and kinematic chains to parameterize \cref{eq:ik}. For closed-loop verification in RoboTwin~2.0~\cite{robotwin2}, the compiler maps object trajectories into the workspace via a height-preserving planar similarity transform, scores $\mathbf{T}_{OG}$ across six closure-axis rolls, and applies centroid clamping (clamping antipodal grasp centers toward the object support-hull centroid) to generate live object-locked phase waypoints and retain physically passing rollouts for VLA post-training.

\section{Experiments}
\label{sec:exp}

We evaluate Ego2X across contact recovery (\cref{sec:exp-core}), decoupled depth correction (\cref{sec:exp-depth}), cross-morphology retargeting and URDF compilation (\cref{sec:exp-ik}), de-skinned compositing (\cref{sec:exp-render}), and closed-loop simulation verification and VLA post-training (\cref{sec:exp-rt2}).

\subsection{Experimental Setup}
\label{sec:exp-setup}

\noindent\textbf{Benchmarks and protocol.}
We evaluate 3D contact and depth accuracy on HOT3D~\cite{hot3d} across disjoint development, confirmation, held-out, and fresh splits, video conversion across $10$ manipulation categories on EgoDex~\cite{egodex}, and closed-loop execution and VLA post-training in RoboTwin~2.0~\cite{robotwin2}. Comparisons pair clips on HOT3D, takes on EgoDex, and scene seeds in RoboTwin~2.0, reporting $10^4$-resample cluster-bootstrap $95\%$ CIs, Wilcoxon signed-rank tests, and Holm-adjusted exact McNemar tests.

\noindent\textbf{Baselines.}
On EgoDex, Ego2X and official HuRo stage-8~\cite{huro} use monocular RGB EgoHand~\cite{egohand} reconstructions, whereas official Ego2Robot~\cite{ego2robot} uses privileged device-tracked 3D hands. We also evaluate a controlled \emph{wrist-following} ablation (sharing our front-end, wrist targets, and solver without contact terms in \cref{eq:ik}) and antipodal gripper baselines on the identical SDF $\phi_o$. Mesh surface clearance and closed-loop physics evaluate manipulation independently of CORE under shared scene mapping, planning, controllers, and verification checks.

\subsection{Contact Recovery}
\label{sec:exp-core}

\begin{table}[t]
\centering
\scriptsize
\renewcommand{\arraystretch}{0.94}
\setlength{\tabcolsep}{1.8pt}
\caption{\textbf{CORE contact recovery on HOT3D.} Top: cue ablation and field-selective readout under dataset object poses ($0.408$ Mode F1; $7.65{\to}3.84$\,mm on $8$ held-out clips). Bottom: estimated object poses ($^{\dagger}$camera-frame distance and velocity state).}
\label{tab:core_acc}
\begin{tabular*}{\columnwidth}{@{\extracolsep{\fill}}lccccc@{}}
\toprule
Cue / Pose setting & Fr.\ F1$\uparrow$ & Iv@.5$\uparrow$ & tIoU$\uparrow$ & Pt (mm)$\downarrow$ & State$\uparrow$ \\
\midrule
\multicolumn{6}{@{}l}{\emph{Reliable (dataset) object pose: upper bound ($31$ clips)}} \\
Geometric ($G$)                      & 0.847 & 0.565 & 0.826 & 6.69 & 0.701 \\
$+$ Tactile \& ray ($T{+}R$)         & 0.912 & 0.690 & 0.889 & \textbf{1.96} & 0.792 \\
$+$ Bridging ($T{+}R{+}B{+}E$, bridged) & 0.920 & 0.722 & \textbf{0.923} & 7.82 & \textbf{0.841} \\
CORE (field-selective, ours)         & \textbf{0.924} & \textbf{0.741} & 0.915 & 2.39 & \textbf{0.841} \\
\midrule
\multicolumn{6}{@{}l}{\emph{End-to-end estimated object pose ($19$ clips)}} \\
Raw FoundationPose                   & 0.191 & 0.065 & 0.255 & 82.44$^{\dagger}$ & 0.076$^{\dagger}$ \\
$+$ Object depth fusion              & \textbf{0.386} & \textbf{0.209} & \textbf{0.571} & \textbf{52.28}$^{\dagger}$ & \textbf{0.202}$^{\dagger}$ \\
\bottomrule
\end{tabular*}
\vspace{-0.5em}
\end{table}

\begin{figure}[!tb]
  \centering
  \includegraphics[width=0.68\columnwidth]{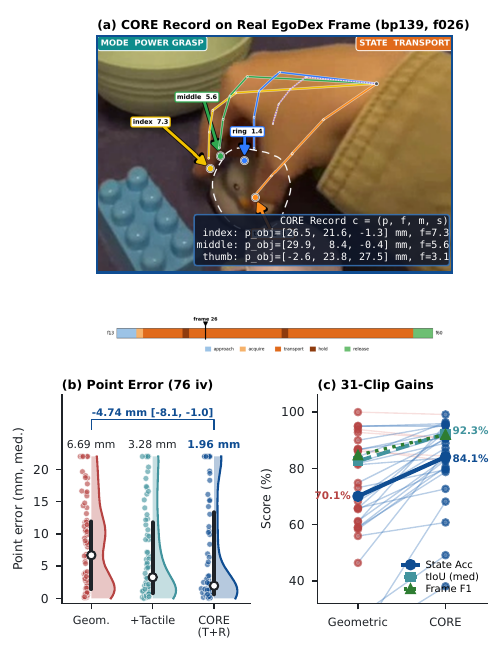}
  \caption{\textbf{Multi-cue CORE recovery under occlusion.} (a) Object-frame anchors and grasp states during an occluded hold. (b,c) Cue ablation ladder on $31$ HOT3D confirmation clips.}
  \label{fig:core}
  \vspace{-0.5em}
\end{figure}

We evaluate CORE against independent HOT3D ground-truth contacts from multi-camera motion-capture MANO-to-object mesh proximity ($\le 5\,\mathrm{mm}$ for $\ge 3$ frames; self-check interval F1 $=1.000$, independent of rays and EgoTac; \cref{tab:core_acc,fig:core}). First, on $31$ confirmation clips with dataset object poses, tactile-seeded ray anchoring ($T{+}R$) lowers 3D anchor error from $6.69\,\mathrm{mm}$ ($G$) to $1.96\,\mathrm{mm}$ ($1.92\,\mathrm{mm}$ held-out). Medoid bridging and state inference ($T{+}R{+}B{+}E$) raise tIoU from $0.826$ to $0.923$ and state accuracy from $0.701$ to $0.841$, while field-selective readout reaches $0.924$ frame F1, $0.741$ interval F1@$0.5$, $0.915$ tIoU, $2.39\,\mathrm{mm}$ error ($7.65\to 3.84\,\mathrm{mm}$ held-out), $0.408$ mode F1, and $0.841$ state accuracy. EgoTac loads reach $0.900$ median Spearman correlation with ground-truth parts ($92.2\%$ within friction cones). Second, under estimated FoundationPose poses ($19$ clips), object depth fusion (\cref{sec:hand-refine}) doubles frame F1 ($0.191$ to $0.386$) and raises tIoU ($0.255$ to $0.571$); the $52.28\,\mathrm{mm}$ camera-frame offset reflects upstream 6-DoF pose error, and tier filtering restricts downstream optimization to \textsc{Certified} and \textsc{Trusted} anchors.

\subsection{Decoupled Depth Correction}
\label{sec:exp-depth}

\begin{table}[t]
\centering
\scriptsize
\renewcommand{\arraystretch}{0.94}
\caption{\textbf{Decoupled depth correction and pose credibility on HOT3D.} Top: GT-free credibility AUROC (calibrated / out-of-fold on LOCO). Middle: hand refinement (mm; raw/ours on dataset poses, $\Delta$ and valid/all coverage on estimated poses). Bottom: in-contact object translation error (mm).}
\label{tab:depth}
\setlength{\tabcolsep}{2.2pt}
\begin{tabular*}{\columnwidth}{@{\extracolsep{\fill}}lcccc@{}}
\toprule
Pose credibility (AUROC $\uparrow$) & LOCO ($36$) & Fresh ($11$) & Conf ($8$) & Held ($11$) \\
\midrule
Single-feature 2D mask IoU  & 0.727 & 0.735 & 0.663 & \textbf{0.888} \\
29-feature monotonic GBDT   & \textbf{0.893} / \textbf{0.886} & \textbf{0.899} & \textbf{0.789} & 0.706 \\
\bottomrule
\end{tabular*}\par
\vspace{0.2em}
\setlength{\tabcolsep}{1.9pt}
\begin{tabular*}{\columnwidth}{@{\extracolsep{\fill}}llcccc@{}}
\toprule
Hand path & Pose & MPJPE $\downarrow$ & Ray $\downarrow$ & To-obj.\ $\downarrow$ & Band $\uparrow$ \\
\midrule
EgoHand ($10$) & data & 54.85/\textbf{48.62} & 51.71/\textbf{44.41} & 26.51/\textbf{21.73} & 25.3/\textbf{33.0\%} \\
HaWoR ($6$)    & data & 19.16/\textbf{15.77} & 14.35/\textbf{9.51} & 12.37/\textbf{9.14} & 56.8/\textbf{70.7\%} \\
\midrule
\multicolumn{2}{@{}l}{Decoupled, est.\ pose} & \multicolumn{2}{c}{MPJPE $\Delta$ [95\% CI]} & To-obj.\ $\Delta$ & Cov.\ val./all \\
\multicolumn{2}{@{}l}{\quad Conf ($8$ clips)} & \multicolumn{2}{c}{$\mathbf{-4.31}$ $[-7.87,-1.38]$} & $\mathbf{-4.52}$ & 97.1/60.9\% \\
\multicolumn{2}{@{}l}{\quad Held ($11$ clips)} & \multicolumn{2}{c}{$\mathbf{-1.91}$ $[-4.53,-0.04]$} & $\mathbf{-2.90}$ & 67.6/35.2\% \\
\bottomrule
\end{tabular*}\par
\vspace{0.2em}
\setlength{\tabcolsep}{3.0pt}
\begin{tabular*}{\columnwidth}{@{\extracolsep{\fill}}lccc@{}}
\toprule
In-contact obj.\ trans.\ (mm) $\downarrow$ & Raw med. & Ours med. & Mean $\Delta$ [95\% CI] \\
\midrule
Confirmation ($8$ clips) & 104.6 & \textbf{67.3} & $-30.4$ $[-44.4, -16.5]$ \\
Held-out ($11$ clips)    & 141.9 & \textbf{99.1} & $-47.3$ $[-78.0, -12.9]$ \\
\bottomrule
\end{tabular*}
\vspace{-0.5em}
\end{table}

\begin{figure}[!tb]
  \centering
  \makebox[0.68\columnwidth][r]{\includegraphics[width=\dimexpr 0.68\columnwidth*26084/23760\relax]{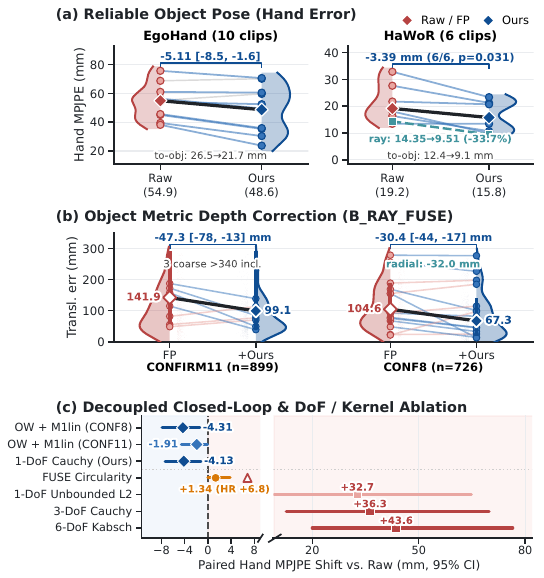}}
  \caption{\textbf{Decoupled 1-DoF Cauchy ray correction on HOT3D.} (a) Hand MPJPE, ray, and surface error. (b) Object translation (\texttt{B\_RAY\_FUSE}; \texttt{CONF8}/\texttt{CONFIRM11}: Conf/Held). (c) Coupled (\texttt{FUSE}) and higher-DoF fits diverge; decoupled \texttt{OW+M1lin} stays stable.}
  \label{fig:depth}
  \vspace{-0.5em}
\end{figure}

\noindent\textbf{Pose credibility.}
For GT-free detection of usable object poses (\cref{tab:depth}, top), our isotonic-calibrated 29-feature monotonic GBDT reaches $0.893$ LOCO AUROC ($0.886$ out-of-fold), $0.899$ on fresh, $0.789$ on confirmation, and $0.706$ on held-out clips, outperforming single-feature 2D mask IoU ($0.727$, $0.735$, $0.663$, and $0.888$) on three splits, while hard single-frame gating retains at most $9.2\%$ valid-slot coverage.

\noindent\textbf{Decoupled hand and object paths.}
With dataset object poses (\cref{tab:depth}, middle; \cref{fig:depth}a), 1-DoF Cauchy ray correction on EgoHand~\cite{egohand} ($10$ clips) lowers contact MPJPE from $54.85$ to $48.62\,\mathrm{mm}$ ($-5.11\,\mathrm{mm}$ paired, $p=0.027$), ray error from $51.71$ to $44.41\,\mathrm{mm}$ ($-14.1\%$), and surface distance from $26.51$ to $21.73\,\mathrm{mm}$ ($-3.52\,\mathrm{mm}$). On HaWoR~\cite{hawor} ($6$ clips), MPJPE drops from $19.16$ to $15.77\,\mathrm{mm}$ ($6/6$ clips, $p=0.031$) and ray error drops by $33.7\%$ ($14.35$ to $9.51\,\mathrm{mm}$). Under estimated poses on confirmation ($8$ clips), object-relative refinement lowers hand MPJPE from $48.6$ to $42.8\,\mathrm{mm}$ ($-5.8\,\mathrm{mm}$, ray $44.7\to 36.8\,\mathrm{mm}$; $-15.4\,\mathrm{mm}$ when pose error is ${<}30\,\mathrm{mm}, {<}15^\circ$), whereas C2Dex-style~\cite{c2dex} absolute camera-ray shift adds $+5.0\,\mathrm{mm}$ relative error and unbounded $L_2$/3-DoF/6-DoF fits diverge ($+32.7$ to $+43.6\,\mathrm{mm}$; \cref{fig:depth}c). Automated interval-propagated triangulation lowers all-slot hand MPJPE by $4.31\,\mathrm{mm}$ ($-4.52\,\mathrm{mm}$ to surface, $97.1/60.9\%$ valid/all coverage) on confirmation and $1.91\,\mathrm{mm}$ ($-2.90\,\mathrm{mm}$, $67.6/35.2\%$) on $11$ held-out clips, avoiding the $+1.34\,\mathrm{mm}$ drift of hand-scaled feedback ($+6.83\,\mathrm{mm}$ hand-only ruler). Meanwhile, the object path cuts median in-contact translation error from $104.6$ to $67.3\,\mathrm{mm}$ (mean $-30.4\,\mathrm{mm}$) and $141.9$ to $99.1\,\mathrm{mm}$ (mean $-47.3\,\mathrm{mm}$; \cref{tab:depth}, bottom; \cref{fig:depth}b).

\subsection{Contact-Driven Retargeting and Compilation}
\label{sec:exp-ik}

\begin{table}[t]
\centering
\scriptsize
\renewcommand{\arraystretch}{0.94}
\setlength{\tabcolsep}{1.8pt}
\caption{\textbf{Contact-driven inverse kinematics for dexterous hands.} Top: HOT3D ($31$ clips; Band: $[-3,+5]$\,mm; Pen.: per-slot / frame union ${<}-3$\,mm). Bottom: EgoDex target-contact realization ($[-5,+10]$\,mm; subsets defined in text).}
\label{tab:ik_dex}
\begin{tabular*}{\columnwidth}{@{\extracolsep{\fill}}lcccc@{}}
\toprule
HOT3D ($31$ clips, $72$ pairs) & Anchor$\downarrow$ & Band$\uparrow$ & Pen.\ (slot/union)$\downarrow$ & $\dot{q}{>}6$ (\%)$\downarrow$ \\
\midrule
Wrist following (\texttt{wf})     & 15.85 & 0.507 & \textbf{0.080} / \textbf{0.353} & \textbf{2.88} \\
Ours, no regularizers             & \textbf{4.97} & \textbf{0.983} & 0.139 / 0.866 & 2.91 \\
Ours, no $E_{\mathrm{temp}}$      & 6.61 & \textbf{0.983} & 0.131 / 0.824 & 17.62 \\
Ours (\texttt{CARD}, Eq.~\ref{eq:ik}) & 6.78 & 0.969 & 0.133 / 0.757 & 3.55 \\
\bottomrule
\end{tabular*}\par
\vspace{0.25em}
\setlength{\tabcolsep}{2.4pt}
\begin{tabular*}{\columnwidth}{@{\extracolsep{\fill}}lccccc@{}}
\toprule
 & \multicolumn{3}{c}{Contact realization $\uparrow$} & \multicolumn{2}{c}{Anchor (mm) $\downarrow$} \\
\cmidrule(lr){2-4}\cmidrule(lr){5-6}
EgoDex ($79$ takes) & ALL & GEOM & GEOM$\cap$HC & ALL & GEOM$\cap$HC \\
\midrule
Human hand (MANO)                 & 0.358 & 0.978 & 0.993 & 29.19 & 2.06 \\
HuRo stage-8~\cite{huro}          & 0.279 & 0.690 & 0.708 & 35.18 & 13.43 \\
Ours, wrist following (\texttt{wf}) & 0.296 & 0.702 & 0.717 & 34.66 & 13.90 \\
Ours (\texttt{CARD}, Eq.~\ref{eq:ik}) & \textbf{0.629} & \textbf{0.876} & \textbf{0.895} & \textbf{8.86} & \textbf{4.52} \\
\bottomrule
\end{tabular*}
\vspace{-0.5em}
\end{table}

\begin{table}[t]
\centering
\scriptsize
\renewcommand{\arraystretch}{0.94}
\caption{\textbf{Parallel-gripper grasp synthesis and cross-embodiment compilation.} Top: true-mesh jaw fit (Band~A: $[-4.5,+5.5]$\,mm of $\phi_o$), far-pad clearance, and coverage on $28$ EgoDex sides ($1{,}272$ frames) and HOT3D. Bottom: five grippers compiled from shared CORE records ($14$ eps) and rate-retimed ALLEX ($81$ eps).}
\label{tab:ik_grip}
\setlength{\tabcolsep}{2.1pt}
\begin{tabular*}{\columnwidth}{@{\extracolsep{\fill}}lcccc@{}}
\toprule
Grasp synthesis ($28$ sides) & Band A$\uparrow$ & Far (mm)$\downarrow$ & Cov.$\uparrow$ & HOT3D$\uparrow$ \\
\midrule
Ego2Robot rule~\cite{ego2robot}       & 0.097 & n/a & 1.00 & 0.648 \\
Wrist / Pair / Tip midpoint           & .006/.085/.114 & 90.3/34.0/32.6 & 1.00 & n/a \\
Per-frame antipodal (no lock)         & 0.454 & 3.34 & 0.75 & n/a \\
Object bbox antipodal ($C_{\mathrm{geom}}$) & 0.499 & 3.27 & 0.75 & n/a \\
CORE grasp ($C_{\mathrm{full}}$, ours) & \textbf{0.632} & \textbf{0.82} & \textbf{1.00} & \textbf{0.998} \\
\bottomrule
\end{tabular*}\par
\vspace{0.25em}
\setlength{\tabcolsep}{1.8pt}
\begin{tabular*}{\columnwidth}{@{\extracolsep{\fill}}lcccc@{}}
\toprule
Compiled embodiment & EEF Real.\ $\uparrow$ & 95\% CI & Pen.\ $\downarrow$ & Arm IK / Free \\
\midrule
Panda + 2F-85        & 0.809 & [0.621, 0.914] & \textbf{0.020} & 0.457 / 0.592 \\
ARX5 (Aloha)         & 0.810 & [0.625, 0.909] & 0.026 & 0.620 / 0.719 \\
Franka Hand          & 0.797 & [0.609, 0.902] & 0.024 & 0.558 / 0.765 \\
UR5 + WSG50          & 0.796 & [0.606, 0.901] & 0.026 & \textbf{0.695} / 0.751 \\
Piper                & \textbf{0.820} & [0.635, 0.918] & 0.027 & 0.032 / \textbf{0.839} \\
ALLEX (rate-retimed) & \textbf{0.902} & [0.846, 0.946] & 0.025 & wrist driven \\
\bottomrule
\end{tabular*}
\vspace{-0.5em}
\end{table}

\begin{figure}[!tb]
  \centering
  \includegraphics[width=0.68\columnwidth]{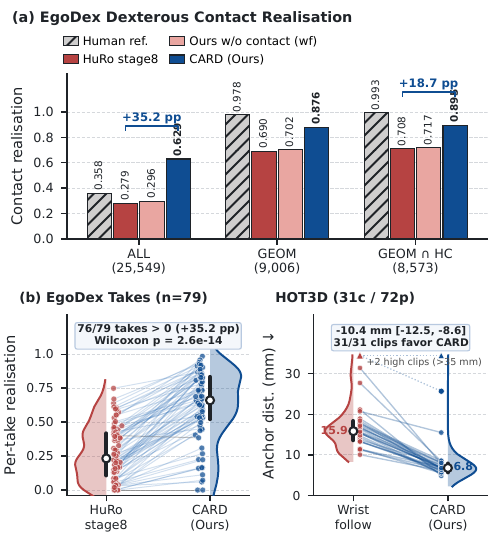}
  \caption{\textbf{Contact-driven IK (\texttt{CARD}, full \cref{eq:ik}) vs.\ wrist following (\texttt{wf}) and HuRo.} (a) Target-contact realization across $79$ EgoDex takes. (b) Pad-to-anchor distance on $31$ HOT3D clips.}
  \label{fig:ikdex}
  \vspace{-0.5em}
\end{figure}

\noindent\textbf{Dexterous hands.}
We evaluate target-contact realization (pad signed distance in $[-5,+10]\,\mathrm{mm}$ on EgoDex and $[-3,+5]\,\mathrm{mm}$ on HOT3D; \cref{tab:ik_dex,fig:ikdex}) across all slots (ALL), slots within $15\,\mathrm{mm}$ of $\phi_o$ (GEOM), and high-confidence EgoTac slots ($\text{GEOM}\cap\text{HC}$). Across $79$ EgoDex takes ($25{,}549$ slots), wrist following (\texttt{wf}) reaches $29.6\%$ (ALL) and $71.7\%$ ($\text{GEOM}\cap\text{HC}$), matching HuRo stage-8 ($27.9\%$ and $70.8\%$). Full \cref{eq:ik} (\texttt{CARD}) reaches $62.9\%$ on ALL ($+35.2\,\mathrm{pp}$ paired over HuRo, $p=2.6\times 10^{-14}$, $76/79$ takes) and $89.5\%$ on $\text{GEOM}\cap\text{HC}$ ($+22.7\,\mathrm{pp}$ on $73$ takes, $p=3.3\times 10^{-10}$; anchor $13.43\to 4.52\,\mathrm{mm}$, vs.\ $99.3\%$ MANO) within $7.5\,\mathrm{rad/s}$ (vs.\ $63.9\,\mathrm{rad/s}$ HuRo), while whole-hand SDF penetration ${<}-1\,\mathrm{mm}$ matches HuRo ($4.4\%$ vs.\ $4.1\%$, $p=0.393$). On $31$ HOT3D clips (\cref{tab:ik_dex}, top), per-slot penetration ${<}-3\,\mathrm{mm}$ is $0.133$ ($0.757$ frame union vs.\ $0.866$ unregularized; $0.353$ for detached \texttt{wf} at $0.507$ band), and $E_{\mathrm{temp}}$ cuts steps ${>}6\,\mathrm{rad/s}$ from $17.6\%$ to $3.6\%$.

\noindent\textbf{Parallel grippers and URDF compilation.}
Measuring 3D mesh jaw fit independently of CORE across $28$ EgoDex sides ($1{,}272$ holding frames; \cref{tab:ik_grip}, top), CORE grasp synthesis reaches $63.2\%$ two-pad fit (Band~A: $[-4.5,+5.5]\,\mathrm{mm}$) and $0.82\,\mathrm{mm}$ far-pad clearance at $100\%$ coverage, outperforming bbox antipodal ($49.9\%$, $3.27\,\mathrm{mm}$, $75\%$; $+13.3\,\mathrm{pp}$), per-frame antipodal ($45.4\%$, $14.1\%$ jumps; $+17.8\,\mathrm{pp}$), and Ego2Robot ($9.7\%$ on EgoDex; $64.8\%$ vs.\ $99.8\%$ on HOT3D). From shared CORE records (\cref{tab:ik_grip}, bottom), five grippers reach $0.796$--$0.820$ EEF realization with $<0.03$ penetration ($0.902$ on ALLEX over $81$ episodes), and fixed table-edge arm IK spans $0.032$ (Piper) to $0.695$ (UR5+WSG50).

\subsection{Contact-Aware Erasure and Compositing}
\label{sec:exp-render}

\begin{table}[!tb]
\centering
\scriptsize
\renewcommand{\arraystretch}{0.94}
\caption{\textbf{Contact-aware erasure and compositing.} Top: held-object pixel loss and residual skin inside the protected region on three audited takes. Bottom: contact-edge skin bleed, balanced part-occlusion accuracy (bPOA), and object vanishing across $14$ unseen EgoDex takes ($663$ frames).}
\label{tab:render}
\setlength{\tabcolsep}{3.2pt}
\begin{tabular*}{\columnwidth}{@{\extracolsep{\fill}}lcc@{}}
\toprule
Erasure mask ($3$ audited takes) & Object pixel loss $\downarrow$ & Skin in region $\downarrow$ \\
\midrule
No object protection & 55.0 / 19.4 / 54.9\% & n/a \\
Object mask only     & n/a & 42.1 / 20.7 / 51.8\% \\
CORE interval (ours) & \textbf{11.2 / 3.0 / 5.2\%} & \textbf{0.03 / 0.34 / 0.31\%} \\
\bottomrule
\end{tabular*}\par
\vspace{0.2em}
\setlength{\tabcolsep}{3.0pt}
\begin{tabular*}{\columnwidth}{@{\extracolsep{\fill}}lccc@{}}
\toprule
Compositing ($14$ unseen takes, $663$ frames) & Skin bleed $\downarrow$ & bPOA $\uparrow$ & Vanish $\downarrow$ \\
\midrule
Contact-protected plate      & 12.55\% & 0.500 & \textbf{0.00\%} \\
Soft depth margin ($10$\,mm) & 8.70\% & 0.566 & \textbf{0.00\%} \\
De-skinned contact (ours)    & \textbf{0.00\%} & \textbf{0.608} & \textbf{0.00\%} \\
\bottomrule
\end{tabular*}
\vspace{-0.5em}
\end{table}

\begin{figure}[!tb]
  \centering
  \includegraphics[width=0.68\columnwidth]{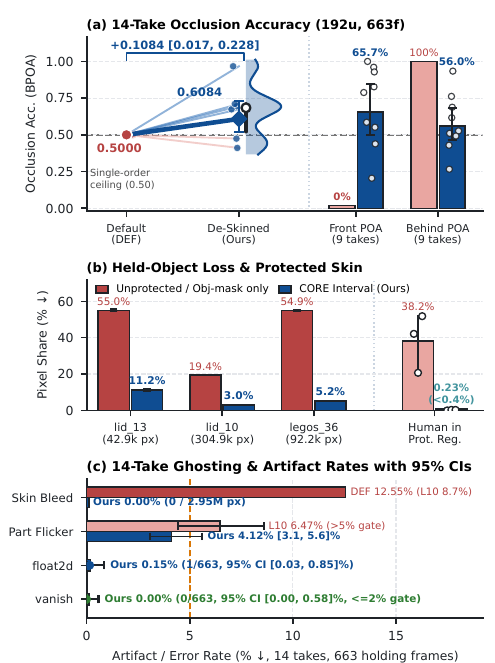}
  \caption{\textbf{Contact-aware erasure and Lab de-skinning.} (a,c) De-skinning across $14$ unseen EgoDex takes. (b) Held-object pixel retention with CORE interval protection.}
  \label{fig:render}
  \vspace{-0.5em}
\end{figure}

Without object protection, arm inpainting erases $19.4\%$--$55.0\%$ of visible held-object pixels on three pixel-audited takes, whereas CORE interval protection reduces this loss to $3.0\%$--$11.2\%$ with at most $0.34\%$ residual skin (\cref{tab:render,fig:render}). Across $14$ unseen EgoDex takes ($663$ frames, $2.95$\,M object pixels), Lab rim de-skinning cuts contact-edge skin bleed from $12.55\%$ to $0.00\%$ and raises balanced part-occlusion accuracy (bPOA, the mean of front-of-object and behind-object occlusion accuracy) from $0.500$ to $0.608$ ($+10.8\,\mathrm{pp}$) with $0.00\%$ object vanishing.

\subsection{Simulation Verification and Policy Learning}
\label{sec:exp-rt2}

\begin{table}[!tb]
\centering
\scriptsize
\caption{\textbf{Closed-loop demonstration yield and X-VLA policy post-training in RoboTwin~2.0.} Top: physics-verified conversion yield on matched videos. Bottom: X-VLA success rate (SR) over $200$ paired scenes with zero simulator demos unless noted (Holm-adjusted exact McNemar $p$ vs.\ zero-shot).}
\label{tab:policy}
\setlength{\tabcolsep}{1.3pt}
\renewcommand{\arraystretch}{0.94}
\begin{tabular*}{\columnwidth}{@{\extracolsep{\fill}}lrrl@{}}
\toprule
Task / Training data & Eps & Yield/SR$\uparrow$ & Comparison, Holm $p$ \\
\midrule
\multicolumn{4}{@{}l}{\emph{Closed-loop demonstration yield (physics verification)}} \\
Bread in basket ($52$ matched)   & 31/52 & \textbf{59.6\%} & vs.\ 1.9\% HuRo/E2R \\
Bell on scale ($53$ specs)       & 45/53 & \textbf{84.9\%} & vs.\ 0.0\%, $4.8{\times}10^{-7}$ \\
\midrule
\multicolumn{4}{@{}l}{\emph{X-VLA: Place bread in basket ($200$ paired scenes)}} \\
Zero-shot X-VLA                  & 0     & 40.5\% & baseline \\
20 simulator expert demos        & 20    & 72.0\% & sim expert ref. \\
HuRo / Ego2Robot$^{*}$           & 1/1   & 2.0/4.0\% & $0{:}29$/$1{:}28$ vs.\ ours \\
Ego2X, unfiltered / $+3$ failed  & 55/73 & 41.0/44.5\% & n.s.\ vs.\ zero-shot \\
Ego2X, verified (\texttt{P70})   & 70    & 63.5\% & $67{:}24$, $7.4{\times}10^{-6}$ \\
Ego2X, verified (\texttt{55 lw}) & 55    & \textbf{67.0\%} & $70{:}21$, $5.0{\times}10^{-7}$ \\
Co-train ($20$ sim $+$ $24$ ego) & 20+24 & \multicolumn{2}{l@{}}{$+5.5$\,pp $[+0.7,+10.3]$ vs.\ HuRo (3 seeds)} \\
\midrule
\multicolumn{4}{@{}l}{\emph{X-VLA: Place bell on scale ($200$ scenes)}} \\
Zero-shot X-VLA                  & 0     & 17.0\% & baseline \\
Ego2X, unfiltered                & 40    & 27.0\% & $17{:}58$, $8.7{\times}10^{-6}$ \\
Ego2X, zero-retry (fixed)        & 20/40 & 47.0/48.0\% & $68{:}8$, $5.6{\times}10^{-13\,{\dagger}}$ \\
Ego2X, zero-retry $+$ prompt     & 40    & \textbf{55.0\%} & $84{:}11$, $2.7{\times}10^{-14}$ \\
\bottomrule
\end{tabular*}
\\[0.15em]\parbox{\columnwidth}{\scriptsize $^{*}$On $50$ scenes ($44.0\%$ zero-shot): $52$-ep unfiltered SR is $60.0\%$ Ours vs.\ $0.0\%$ HuRo / $2.0\%$ E2R ($p{=}3.7{\times}10^{-9}$); verified SR is $70.0\%$. $^{\dagger}$On $195$ paired scenes.}
\vspace{-0.5em}
\end{table}

\begin{figure}[!tb]
  \centering
  \includegraphics[width=0.68\columnwidth]{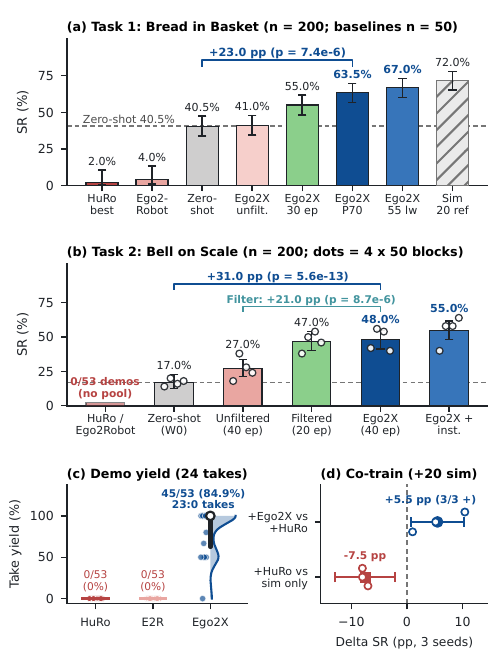}
  \caption{\textbf{Closed-loop verification and X-VLA policy learning.} (a) Place bread in basket over $200$ paired scenes (\texttt{P70}: $70$ verified; \texttt{55 lw}: $55$ layout-weighted). (b) Place bell on scale ablations. (c) Matched demonstration yield. (d) Co-training with simulator demos.}
  \label{fig:policy}
  \vspace{-0.5em}
\end{figure}

\noindent\textbf{Executable demonstration yield.}
Under identical simulator checks on matched RoboTwin~2.0 videos (\cref{fig:teaser}c, \cref{tab:policy}, \cref{fig:policy}c), Ego2X converts $31/52$ ($59.6\%$) \emph{Place bread in basket} videos vs.\ $1/52$ ($1.9\%$) for HuRo and Ego2Robot, and reaches $38.8\%$ phase-1 feasibility across $348$ layouts vs.\ $29.0\%$ for Ego2Robot (Holm $p=0.019$) and $35.6\%$ for HuRo. On $53$ \emph{Place bell on scale} specs ($24$ takes), Ego2X yields $45/53$ ($84.9\%$; $40/53=75.5\%$ zero-retry) vs.\ $0/53$ for both baselines (Holm $p=4.8\times10^{-7}$). Removing $\mathbf{T}_{OG}$ binding drops yield to $0/53$ (lifting $52/53$ before failing placement, $p=3.4\times10^{-13}$), removing CORE state bounds yields $39/53$, and removing centroid clamping drops anchor-free grasps from $48/53$ to $8/53$ ($p=2.0\times10^{-11}$).

\noindent\textbf{VLA post-training.}
Post-training X-VLA~\cite{xvla} on verified Ego2X episodes alone raises \emph{Place bread in basket} success rate over $200$ paired scenes from $40.5\%$ zero-shot to $67.0\%$ ($55$ layout-weighted episodes, \texttt{55 lw}, Holm $p=5.0\times10^{-7}$; \cref{fig:teaser}c) and $63.5\%$ ($70$ episodes, \texttt{P70}, $p=7.4\times10^{-6}$; vs.\ $72.0\%$ for $20$ simulator demos, \cref{fig:policy}a), whereas unfiltered pools ($41.0\%$, $44.5\%$) stay near zero-shot. On the initial $50$ scenes ($44.0\%$ zero-shot), $52$-episode unfiltered training reaches $60.0\%$ vs.\ $0.0\%$ (HuRo) and $2.0\%$ (Ego2Robot, $p=3.7\times10^{-9}$), and verified Ego2X reaches $70.0\%$ vs.\ $2.0\%$ and $4.0\%$. In 3-seed co-training ($20$ simulator $+$ $24$ unfiltered episodes; \cref{fig:policy}d), HuRo lowers simulator-only success rate by $-7.5\,\mathrm{pp}$, whereas Ego2X beats HuRo by $+5.5\,\mathrm{pp}$ ($[+0.7,+10.3]$) on all seeds. On \emph{Place bell on scale} ($200$ scenes, $17.0\%$ zero-shot; \cref{fig:policy}b), $40$ zero-retry episodes reach $55.0\%$ with task-aligned prompts ($p=2.7\times10^{-14}$) and $48.0\%$ with a fixed placeholder prompt ($p=5.6\times10^{-13}$ on $195$ scenes; $47.0\%$ at $20$ episodes), vs.\ $27.0\%$ unfiltered ($p=8.7\times10^{-6}$).

\section{Conclusion}
\label{sec:conclusion}

We presented Ego2X, an object-centric framework that converts monocular egocentric video into physically executable robot demonstrations. Specifically, CORE records object-frame surface anchors, two-part force priors, grasp modes, and interaction states. This unified representation enables decoupled camera-ray depth correction, cross-morphology retargeting, de-skinned compositing, URDF compilation, and closed-loop simulation verification across dexterous hands and parallel grippers. To the best of our knowledge, this paper is the first to explore post-training embodied models using egocentric videos. We hope this work will open a new avenue for scalable post-training of embodied models by unlocking egocentric human videos. With CORE, Ego2X improves the success rate of the X-VLA policy by 26.5\% without relying on expert simulation demonstrations.

\paragraph{Future Work.}
We plan to extend Ego2X in three directions. First, contact point localization remains limited by object pose estimation errors under heavy occlusion. Second, CORE models contact loads as visual-kinematic priors rather than sensor-calibrated Newton forces, which we aim to address via tactile integration. Finally, due to current hardware and schedule constraints, physical robot deployment is still underway and will be completed in future work.

{
    \small
    \bibliographystyle{ieeenat_fullname}
    \bibliography{main}
}


\end{document}